\documentclass[sigconf]{aamas} 

\usepackage{balance} % for balancing columns on the final page

\usepackage{bbm}
\usepackage{macros}

\setcopyright{ifaamas}
\acmConference[AAMAS '26]{Proc.\@ of the 25th International Conference
on Autonomous Agents and Multiagent Systems (AAMAS 2026)}{May 25 -- 29, 2026}
{Paphos, Cyprus}{C.~Amato, L.~Dennis, V.~Mascardi, J.~Thangarajah (eds.)}
\copyrightyear{2026}
\acmYear{2026}
\acmDOI{}
\acmPrice{}
\acmISBN{}

\acmSubmissionID{292}

\title[On the Trade-Off Between Transparency and Security in Adversarial Machine Learning]{On the Trade-Off Between Transparency and Security in Adversarial Machine Learning}
\author{Lucas Fenaux}
\affiliation{
  \institution{University of Waterloo}
  \city{Waterloo}
  \country{Canada}}
\email{lucas.fenaux@uwaterloo.ca}

\author{Christopher Srinivasa}
\affiliation{
  \institution{Borealis AI}
  \city{Toronto}
  \country{Canada}}
\email{christopher.srinivasa@rbc.com}

\author{Florian Kerschbaum}
\affiliation{
  \institution{University of Waterloo}
  \city{Waterloo}
  \country{Canada}}
\email{florian.kerschbaum@uwaterloo.ca}

\begin{abstract}
Transparency and security are both central to Responsible AI, but they may conflict in adversarial settings. We investigate the strategic effect of transparency for agents through the lens of transferable adversarial example attacks. In transferable adversarial example attacks, attackers maliciously perturb their inputs using surrogate models to fool a defender's target model. These models can be defended or undefended, with both players having to decide which to use. Using a large-scale empirical evaluation of nine attacks across 181 models, we find that attackers are more successful when they match the defender's decision; hence, obscurity could be beneficial to the defender. With game theory, we analyze this trade-off between transparency and security by modeling this problem as both a Nash game and a Stackelberg game, and comparing the expected outcomes. 
Our analysis confirms that only knowing whether a defender's model is defended or not can sometimes be enough to damage its security. This result serves as an indicator of the general trade-off between transparency and security, suggesting that transparency in AI systems can be at odds with security.
Beyond adversarial machine learning, our work illustrates how game-theoretic reasoning can uncover conflicts between transparency and security.

\end{abstract}


\newcommand{\BibTeX}{\rm B\kern-.05em{\sc i\kern-.025em b}\kern-.08em\TeX}

\begin{document}

%%% The following commands remove the headers in your paper. For final 
%%% papers, these will be inserted during the pagination process.

\pagestyle{fancy}
\fancyhead{}

%%% The next command prints the information defined in the preamble.

\maketitle 

%%%%%%%%%%%%%%%%%%%%%%%%%%%%%%%%%%%%%%%%%%%%%%%%%%%%%%%%%%%%%%%%%%%%%%%%
% {\color{red}[NOTE] Perhaps naming the choice of defended/undefended surrogate or target model as a decision by the attacker/defender would make describing using the same defense status clearer and easier. Can be said as same decision, and revealing the defense status as the attacker knowing the defender's decision. Fits better into the game theory angle.}
\section{Introduction}

Responsible AI requires that systems and agents are designed, developed, and deployed so that they remain safe, trustworthy, fair, transparent, and ethically sound throughout their lifecycle \cite{batool2023responsible}. 
Two pillars of Responsible AI, \emph{transparency} (e.g., disclosing that AI is used or explaining model outputs) and \emph{security} (ensuring the safe operation of AI agents in safety-critical settings), can intuitively appear to be at odds. In this paper, we investigate whether a trade-off exists between the objectives of these pillars through large-scale experimentation and game-theoretic analysis.

In AI security, problems can usually be modeled as a game between two agents: the attacker, who tries to maximize its probability of success, and the defender, who attempts to minimize this probability \cite{bose2020adversarial, meunier2021mixed}. This setup can be modeled as a zero-sum game with the success probability or one minus that probability as utility functions for attackers and defenders. 

A particular type of attack, with a long history of investigation and that is deployable in the physical world, is the adversarial example attack \cite{kurakin2018adversarial}. An adversarial example is a small perturbation to an input that is imperceptible to humans but leads to a misclassification by the AI model, for example, a slightly modified image of a dog that is confidently classified as a cat. Transferable adversarial example attacks are a type of adversarial example attack that utilizes a surrogate model to generate the adversarial perturbation before submitting the modified image for inference to the target model. Due to their decade-long study, numerous publicly available models have been trained to withstand attacks with adversarial perturbations and achieve the highest defense success rate \cite{robustbench}. Therefore, these models can be used to represent attempts by the defender to secure their model. Additionally, their extensive history of investigation has led to repositories of transferable adversarial example attacks that we can use to represent the attacker \cite{transferattack}. Using both these defenses and transferable adversarial example attacks, we can study the trade-off between transparency and security by constructing security games.
We can then analyze versions of these security games where the defender is transparent and compare them to versions where the defender is not transparent to determine whether there is a trade-off between transparency and security. The general dilemma between security and transparency encompasses many aspects of AI systems throughout their pipeline, whether it be the code, model weights, or defenses \cite{papernot2018sok}. 

In this paper, we focus our efforts on transferable adversarial example attacks, rather than adversarial example attacks in general, for two key reasons. First, the attacker only needs one inference attempt for a successful attack; hence, defending the model itself seems to be the most promising defense strategy. In contrast, white-box attacks can be prevented by keeping the model secret, and black-box attacks can be mitigated by rate limiting. Second, defense transparency is an aspect of transparency that companies like Anthropic \cite{anthropic2025constitutional, anthropic2025claudechrome} and OpenAI \cite{openai2025traderobustness} have been championing. Defense transparency is clearly evident in transferable adversarial example attacks, where the defender can choose to disclose whether they defended their model, and the attacker can adapt to this decision by either defending their surrogate model or not. In particular, the attacker's capability to adapt is driven by the observation that an attacker matching the defender's decision can improve the attack's success rate. This observation is derived from our large-scale empirical analysis of transferable adversarial example attacks and defenses, which spans nine attacks and 181 models. With it, we can study a specific instance of the transparency and security trade-off that focuses on the defender's decision to publicize whether it defended its model or not. In this instance, we show that \textbf{it can be beneficial to security for an AI provider to conceal the fact that they are defending against adversarial example attacks}. 
Our results also support the notion that using a mixture of defenses, when combined with defense obscurity, can further enhance an AI system's robustness against attacks.

Furthermore, our large-scale empirical evaluation also highlights that the potency of transferable adversarial example attacks has been severely underestimated, especially against defended target models. Existing works benchmark attacks using exclusively undefended surrogates \cite{transferattack, wang2021admix, gubri2022lgv, wang2021enhancing, luo2022frequency, zhang2022beyond, guo2025boosting, ge2023boosting, naseer2019cross}, which, our analysis shows, can underestimate the accuracy degradation incurred by attacks by up to $\mathbf{3.73\times}$.

In the remainder of the paper, we give some background on the threat of adversarial examples (see Section \ref{sec:background}), review related work (see Section \ref{sec:related_work}), and explain how we choose attacks and defenses and how we evaluate their success rate (see Section \ref{sec:experiments}). Finally, we present our game-theoretic analysis in Section \ref{sec:game-theory} and conclude the paper in Section \ref{sec:conclusion}.

\section{Background}\label{sec:background}
For our study, we consider image classification tasks and their associated adversarial examples. We denote $\mathcal{X} = \mathbb{R}^d$ as the input space of the task and $\mathcal{Y} = \{0, \dots k\}$ as its output space, with $[0, \dots k]$ classification labels. We let $\mathcal{Z}$ be a data-distribution over $\mathcal{X} \times \mathcal{Y}$, from which we can sample input-label pairs $(x,y) \sim \mathcal{Z}$. Machine learning (ML) models $M_\theta: \mathcal{X} \rightarrow \mathcal{Y}$ are parametrized functions with $\theta \in \mathbb{R}^{m}$, with $\mathcal{M}$ the distribution of ML models for a given task. 

\subsection{Adversarial example attacks}\label{subsec:adv_ex_attack_back}
Adversarial example attacks generate a small perturbation that is added to an input, causing the model to misclassify the input.
For classification tasks, an adversarial example attack can be defined as a zero-sum two-player game between an attacker and a defender as follows:

The defender's action is to train a target model $M_\theta$, which can be interpreted as sampling a model $M_\theta \sim \mathcal{M}$ from the distribution of models for a particular classification task.
The attacker's action is to construct an attack $A: \mathcal{M} \times \mathcal{X} \times \mathcal{Y} \rightarrow \mathcal{X}$ to attack the defender's target model $M_\theta$.

The utility function represents a player's preference for an outcome. For the adversarial example game, the defender's utility $u_d$ is defined as the target model's adversarial accuracy:
\begin{equation}
    u_d(M_\theta, A) = \mathbb{E}_{(x,y) \sim \mathcal{Z}}[\mathbbm{1}(M_{\theta}(A(M_\theta, x, y), y)]
\end{equation}
where $\mathbbm{1}$ is the identity function. The attacker's utility is then $u_a(M_\theta, A) = 1-u_d(M_\theta, A)$ since it is a zero-sum game. The attacker's utility corresponds to the accuracy degradation incurred by the attack. 

Additionally, we constrain the output domain of the attack for a given $(x,y)$ input-label pair so that the perturbed image remains visually indistinguishable from a benign image, with its semantic information remaining unchanged. Otherwise, one could use a perturbation of large magnitude that destroys the semantic information in the image, causing a misclassification by the target model. With $\delta = A(x,y) - x$, we formalize this constraint to the attacker as:
\begin{equation}
    \Vert \delta \Vert_{l} \leq \epsilon
\end{equation}
for some distance-metric $l$ and some dataset-specific $\epsilon$ value. For our experiments, we use the $l_\infty$-norm.
Adversarial example misclassifications can also be targeted towards a particular label \cite{wang2023towards}, where success is when $M_{\theta}(x+ \delta) = y_t$ for some label $y_t$. For our experiments, we focus solely on untargeted attacks as they are easier to carry out.

The sensitivity of machine learning models to small changes in their inputs can have a substantial impact on the security of a machine learning system. This impact is particularly pronounced in safety-critical domains, such as healthcare \cite{ml_healthcare}, autonomous vehicles \cite{GUPTA2021100057}, and aerospace and aviation systems \cite{Ni_2022}, where machine learning systems are routinely deployed. Historically, adversarial attacks have been categorized as either digital or physical, based on whether they are meant to be deployed through software on digital images or in the real world through physical modifications. There also exists a finer categorization derived from the level of access to the target model required to carry out the attack \cite{vassilev2024adversarial}:

\noindent \textbf{White-box} attacks use the model parameters $\theta$ to craft their attack \cite{autopgd}. 

\noindent \textbf{Black-box} attacks can query the model $M_{\theta}$ as an oracle without access to the internals of the model or its parameters $\theta$ \cite{andriushchenko2020square}.

\noindent \textbf{Transferable} attacks utilize surrogate models, also known as shadow models, which are typically trained on the same dataset as the target model. They craft their perturbation $\delta$ against these surrogate models, while trying to increase the likelihood that the perturbation will also work against the target model by improving its transferability \cite{gubri2022lgv}.

In the white-box and black-box settings, the attacker has access to the target model's parameters or can query the target model, respectively, meaning that the attacker can react to the defender's action. Hence, both white-box and black-box adversarial example attacks can be represented as Stackelberg games \cite{stackelberg1934marktform}, with the defender serving as the leader and the attacker as the follower.

In practice, white-box attacks are unrealistic against most real-world systems, as model parameters are usually closely guarded secrets. Likewise, black-box attacks can be countered mainly by proper query rate limiting. Transferable attacks do not require query or parameter access to the target model; instead, they only need access to the training data, which is typically composed of publicly available data sources. They only interact with the target model when delivering their payload, making it the most realistic and threatening of the three. Therefore, we focus our study on transferable attacks. 
% Transferable attacks are categorized by the TransferAttack benchmark\cite{transferattack}  based on the method used to generate the adversarial perturbation as follows: gradient-based, input-transformation-based, advanced objective, model-related, ensemble-based, and generation-based. 
In transferable attacks, since the attacker has no access to the target model, both players make their moves simultaneously, without one adapting to the other. Hence, transferable attacks can be modeled as a Nash game rather than a Stackelberg game. 

\subsection{Adversarial example defenses}\label{subsec:adv_ex_defenses}

To counter adversarial example attacks, researchers have developed various defenses \cite{wu2023defenses, vassilev2024adversarial, robustbench}. Defenses are either empirical or provable \cite{vassilev2024adversarial}, each with its own trade-off. Provable defenses, such as randomized smoothing \cite{lecuyer2019certified} or formal verification  \cite{katz2017reluplex}, offer provable guarantees but are often impractical due to their high computational costs. Empirical defenses have more reasonable computational requirements but do not offer provable guarantees. They can be divided into three main approaches \cite{wu2023defenses}: 

\begin{itemize}
    \item \underline{Robust training} \cite{madry2017towards} teaches a model to be robust to adversarial example attacks on top of their main task. Robust training turns model training into a multi-objective optimization problem, balancing robustness and benign accuracy.
    \item  \underline{Adversarial example detection} \cite{xu2017feature} identifies if the input was adversarially manipulated and, if so, rejects it.
    \item \underline{Input purification methods} \cite{nie2022diffusion, shi2021online} sanitize inputs by removing the malicious perturbation added to inputs if it exists.
\end{itemize}

\section{Related Work}\label{sec:related_work}
Game theory has long been used to study security and privacy problems \cite{do2017game, manshaei2013game, pawlick2019game}. A common use case is analyzing the impact of information on an attacker's success. While, in theory, security should not be solely achieved through obscurity, it often proves helpful. A clear example is the white-box vs. black-box vs. transferable adversarial example attack scenario \cite{papernot2018sok}, where the more an attacker knows, the more effective the attack becomes. In practice, Responsible AI practices like transparency explicitly go against obscurity \cite{batool2023responsible}, and therefore can compromise an agent's security. To the best of our knowledge, there has been no research examining how transparency about the model's defense status affects its security.

More recently, game theory has been applied to study adversarial examples, with some works focusing on the white-box and black-box setting \cite{rathbun2022game, araujo2020advocating, pal2020game, sengupta2019mtdeep} and others studying the transferability of adversarial examples \cite{rathbun2022game, sengupta2019mtdeep, bose2020adversarial}. These works and existing surveys \cite{dasgupta2019survey, li2024survey} highlight the role of attacker knowledge and information asymmetry in adversarial settings; following, to some extent, the adversarial example game we describe in Section \ref{subsec:adv_ex_attack_back}. However, they do not examine the decision problem we study: the defender secretly (or not) chooses whether to deploy a defended or undefended target model, and the attacker decides whether to craft adversarial examples using a defended or undefended surrogate. We investigate this decision problem as a new type of adversarial example game, highlighting the trade-off between transparency and security, where it is beneficial for the parties to keep their choices secret.

Adversarially trained models have been found to transfer better to downstream tasks, leading to surrogate refinement as a method to improve the transferability of adversarial examples across models \cite{zhao2023revisiting}. The lack of a proper large-scale study on whether defended models transfer better has led to an underestimation of the real risk posed by transferable adversarial example attacks. Existing benchmarks and evaluations \cite{li2023towards, transferattack} do not consider using defended surrogates, which, as we observe, disfavor transferable attacks against defended models. As a motivating result for their Cascade defense, Na et al. \cite{na2017cascade} observed in an initial evaluation that the I-FGSM attack \cite{kurakin2016adversarial} transferred more effectively between adversarially trained models. However, they did not further study the impact on transferable attack performance; instead, they focused on building a defense based on their observation. Their motivating results were obtained by transferring between models of the same architecture, but with different initializations. They experimented with two model architectures on the CIFAR10 \cite{cifar10} and MNIST \cite{mnist} datasets. We instead demonstrate that, in general, transferable attacks can be transferred between defended models, irrespective of their specifics. We do so with a large-scale experiment in Section \ref{sec:experiments}. We then analyze our results using game theory to conclude that transparency can be at odds with security in Responsible AI.

\section{Experiments}\label{sec:experiments}

We conduct a large-scale study of transferable adversarial example attacks on image classification models. We choose image classification models because they have been a primary focus of adversarial example research over the past decade, with well-established benchmarks and a large number of attacks and defenses publicly available \cite{transferattack, robustbench}. In particular, the CIFAR-10 \cite{cifar10} and ImageNet \cite{Imagenet} datasets have been extensively studied, with benchmarks such as RobustBench \cite{robustbench} compiling and comparing defenses, as well as making the defended models publicly available. 
The results we present in Section \ref{subsec:evaluation} motivate our defense decision problem. We also use these results as the foundation for our game-theoretic analysis in Section \ref{sec:game-theory}.

\begin{table}[t]
  \centering
  \caption{Average model benign accuracy (in \%) and the number of models averaged per dataset and per setting.}
  \label{tab:baseline}
  % \begin{adjustbox}{width=0.95\columnwidth}

\begin{tabular}{|l|cc|cc|}
\toprule
 & \multicolumn{2}{|c|}{Undefended} & \multicolumn{2}{|c|}{Defended}\\
& Accuracy & Count & Accuracy & Count \\
\midrule
CIFAR-10 & 94.02 & 12 & 87.72 & 66 \\
ImageNet & 75.78 & 68 & 71.66 & 17 \\
\bottomrule
\end{tabular}

% \end{adjustbox}
\end{table}

\subsection{Setup}\label{subsec:setup}

\noindent \textbf{CIFAR10} \cite{cifar10}: The training set contains 50,000 images, and the validation set contains 10,000 images. Both sets are labeled, spanning 10 classes. The images in the CIFAR10 dataset all have a resolution of 32x32 pixels (and the 3 RGB channels). We evaluate each attack on the entire testing set in all of our experiments. \\

\noindent \textbf{ImageNet} \cite{Imagenet}: The training and validation sets consist of 1,281,167 and 50,000 labeled images distributed across 1,000 classes, respectively. The images vary in resolution and quality, reflecting the real-world diversity. They are often pre-processed and resized to a resolution of 224x224 pixels (with 3 RGB channels) for models like ResNet \cite{resnet}. In all our experiments, we assess each attack on the 5,000-image subset of the testing set used by the RobustBench \cite{robustbench} framework in its evaluations. \\

\noindent \textbf{Models: }\label{par:model_provenance} 
We assemble a total of 98 undefended models and 92 defended models to benchmark the attacks. Out of the 98 undefended models, 17 are CIFAR10 models from the pytorch-cifar GitHub repository \cite{pytorch-cifar}. The remaining 81 undefended models are Pytorch's \cite{Pytorch} pre-trained undefended ImageNet models. The defended models are from the RobustBench \cite{robustbench} framework, comprising 71 CIFAR10 models and 21 ImageNet models. These models are compiled from a large corpus of academic research papers that span the various categories of defenses we highlighted in Section \ref{subsec:adv_ex_defenses}. Out of the 190 models, nine do not work with at least one of the attacks, so we exclude them from our evaluation. More specifically, certain model architectures are specialized for inference, which conflicts with the re-training necessary for the LGV attack \cite{gubri2022lgv}. The remaining models that do not work caused memory issues with at least one of the attacks. Out of the working models, 18 are used as surrogates for the transferable attacks, as follows: we use four undefended models and three defended models for CIFAR10, and eight undefended models and three defended models for ImageNet. To avoid evaluating surrogate models against themselves, which would constitute a white-box attack, we also exclude them from the set of target models for our evaluation. 
Hence, for every experiment we conduct, we evaluate each attack against 12 undefended and 66 defended CIFAR10 models, as well as 68 undefended and 17 defended ImageNet models. We present their average benign accuracy on the CIFAR10 and ImageNet validation sets in Table \ref{tab:baseline}.
RobustBench uses an $\epsilon$ of $4/255$ for ImageNet and $8/255$ for CIFAR10; therefore, we use the same $\epsilon$ values when running attacks.
\\

\noindent \textbf{Attacks:}
To perform our evaluation, we gather a variety of transferable attacks: Admix \cite{wang2021admix}, VNIFGSM \cite{wang2021enhancing}, LGV \cite{gubri2022lgv}, SSAH \cite{luo2022frequency}, BIA \cite{zhang2022beyond}, OPS \cite{guo2025boosting}, PGN \cite{ge2023boosting}, CDTP \cite{naseer2019cross}, and AutoAttack \cite{autopgd}. 
We include AutoAttack \cite{autopgd} as a transferable attack (by attacking a surrogate model) because it is the attack employed by RobustBench for its benchmark. 
We select this subset of attacks from the literature using three selection factors: the attack uses the $l_\infty$-norm for $\epsilon$, the attack has a working publicly available code implementation, and the final set of attacks covers the types of transferable attacks categorized by TransferAttack benchmark \cite{transferattack}: gradient-based, input-transformation-based, advanced objective, model-related, ensemble-based, and generation-based. 
Since our study encompasses a wide range of models, we do not include model-specific attacks in our evaluation. However, we cover the remaining attack types with at least one attack each.

\subsection{Evaluation}\label{subsec:evaluation}
\begin{figure}[t]
    \centering
    \includegraphics[width=0.8\columnwidth]{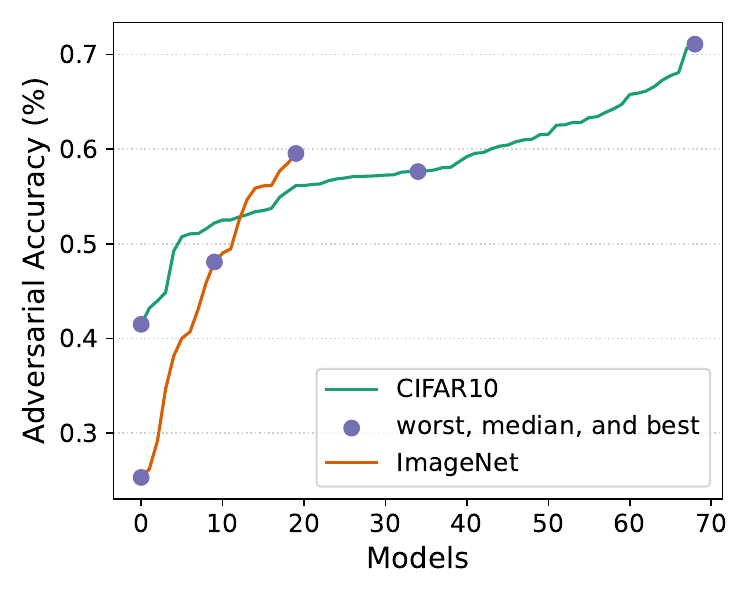}
    \caption{Adversarial accuracy against white-box AutoAttack of the 71 CIFAR10 and the 21 ImageNet defended models, with the worst, median, and best performing models highlighted.}
    \label{fig:small_adv_acc}
\end{figure}

In Table \ref{tab:main-results-cifar10-w-base} and Table \ref{tab:main-results-imagenet-w-base}, we present the average accuracy degradation caused by each attack on CIFAR10 and ImageNet against all the models we evaluate, separated as either undefended or defended target models. We use four types of surrogates for each attack. In the tables, "Und." stands for undefended models, where we either use the models explicitly specified in the original papers, if available, or a fixed set of default pre-trained surrogate models. Ideally, to study defended models as surrogates, we would prefer to use each defended model as a surrogate model and evaluate it against every other model. However, that is not computationally feasible. For that reason, we rank all our defended models per dataset based on their adversarial accuracy against AutoAttack in the white-box setting to estimate their robustness, following the RobustBench benchmark \cite{robustbench}. We then select the worst, the median, and the best performers as our surrogate models, which are labeled as "Worst-Def", "Median-Def", and "Best-Def", respectively. In Section \ref{subsec:surrogate_importance}, we demonstrate that the robustness of the defended surrogate matters little.
The models selected for CIFAR10 and ImageNet are highlighted in Figure \ref{fig:small_adv_acc}, along with the overall adversarial accuracy against white-box AutoAttack of all the defended models we gathered. \\

% [NOTE: If running out of room, can remove this figure and the sentence where it is mentioned.]

\begin{table*}[t]
\centering
\caption{Accuracy degradation (in \%) caused by transferable adversarial attacks against defended and undefended models using undefended, worst-defended, median-defended, and best-defended surrogates for CIFAR10. We include the difference with the No Attack baseline in parentheses.} %We also include white-box AutoAttack \cite{autopgd} and the black-box attack Square \cite{andriushchenko2020square} for comparative purposes.
\label{tab:main-results-cifar10-w-base}
\begin{adjustbox}{width=0.99\linewidth}
\begin{tabular}{lcccccccc}
\toprule
Target & \multicolumn{4}{c}{Undefended} & \multicolumn{4}{c}{Defended} \\
\cmidrule(lr){2-5}
\cmidrule(lr){6-9}
Surrogate & Und. & Worst-Def & Median-Def & Best-Def & Und. & Worst-Def & Median-Def & Best-Def \\
\midrule
No Attack & \multicolumn{4}{c}{5.98} & \multicolumn{4}{c}{12.28} \\
Admix & 96.26 (+90.28) & 24.67 (+18.69) & 25.84 (+19.86) & 31.92 (+25.94) & 16.53 (+4.25) & \textbf{24.00 (+11.72)} & 27.84 (+15.56) & 27.79 (+15.51) \\
AutoAttack & 96.57 (+90.59) & 16.22 (+10.24) & 16.55 (+10.57) & 19.67 (+13.69) & 13.87 (+1.59) & 19.17 (+6.89) & 25.05 (+12.77) & 23.92 (+11.64) \\
BIA & 45.88 (+39.90) & 7.63 (+1.65) & 15.38 (+9.40) & 9.22 (+3.24) & 13.28 (+1.00) & 13.46 (+1.18) & 13.11 (+0.83) & 12.95 (+0.67) \\
CDTP & 81.76 (+75.78) & 12.31 (+6.33) & 7.58 (+1.60) & 7.66 (+1.68) & 13.31 (+1.03) & 15.20 (+2.92) & 13.93 (+1.65) & 14.71 (+2.43) \\
LGV & \textbf{98.43 (+92.45)} & \textbf{52.00 (+46.02)} & \textbf{91.89 (+85.91)} & \textbf{95.48 (+89.50)} & 15.34 (+3.06) & 21.56 (+9.28) & 18.65 (+6.37) & 16.08 (+3.80) \\
OPS & 87.59 (+81.61) & 16.16 (+10.18) & 15.53 (+9.55) & 19.47 (+13.49) & \textbf{16.89 (+4.61)} & 21.40 (+9.12) & 22.40 (+10.12) & 22.92 (+10.64) \\
PGN & 95.40 (+89.42) & 19.35 (+13.37) & 25.50 (+19.52) & 26.67 (+20.69) & 16.02 (+3.74) & 22.89 (+10.61) & 28.30 (+16.02) & 25.83 (+13.55) \\
SSAH & 9.26 (+3.28) & 15.68 (+9.70) & 13.42 (+7.44) & 11.06 (+5.08) & 12.42 (+0.14) & 14.58 (+2.30) & 12.80 (+0.52) & 12.87 (+0.59) \\
VNI-FGSM & 85.55 (+79.57) & 22.28 (+16.30) & 29.82 (+23.84) & 32.78 (+26.80) & 15.68 (+3.40) & 22.61 (+10.33) & \textbf{29.47 (+17.19)} & \textbf{27.84 (+15.56)} \\
\midrule
Mean & 77.41 (+71.43) & 20.70 (+14.72) & 26.83 (+20.85) & 28.21 (+22.23) & 14.82 (+2.54) & 19.43 (+7.15) & 21.28 (+9.00) & 20.55 (+8.27) \\
\bottomrule
\end{tabular}
\end{adjustbox}
\end{table*}
\begin{table*}[t]
\centering
\caption{Accuracy degradation (in \%) caused by transferable adversarial attacks against defended and undefended models using undefended, worst-defended, median-defended, and best-defended surrogates for ImageNet. We include the difference with the No Attack baseline in parentheses.}
\label{tab:main-results-imagenet-w-base}
\begin{adjustbox}{width=0.99\linewidth}
\begin{tabular}{lcccccccc}
\toprule
Target & \multicolumn{4}{c}{Undefended} & \multicolumn{4}{c}{Defended} \\
\cmidrule(lr){2-5}
\cmidrule(lr){6-9}
Surrogate & Und. & Worst-Def & Median-Def & Best-Def & Und. & Worst-Def & Median-Def & Best-Def \\
\midrule
No Attack & \multicolumn{4}{c}{24.22} & \multicolumn{4}{c}{28.34} \\
Admix & \textbf{71.09 (+46.87)} & 29.80 (+5.58) & 33.42 (+9.20) & 32.52 (+8.30) & 30.49 (+2.15) & 33.87 (+5.53) & 35.95 (+7.61) & \textbf{36.49 (+8.15)} \\
AutoAttack & 51.23 (+27.01) & 25.86 (+1.64) & 28.68 (+4.46) & 28.09 (+3.87) & 29.26 (+0.92) & 30.47 (+2.13) & 33.09 (+4.75) & 33.22 (+4.88) \\
BIA & 50.54 (+26.32) & 24.82 (+0.60) & 24.60 (+0.38) & 24.26 (+0.04) & 29.29 (+0.95) & 28.99 (+0.65) & 28.14 (-0.20) & 28.24 (-0.10) \\
CDTP & 63.21 (+38.99) & 25.90 (+1.68) & 25.62 (+1.40) & 25.18 (+0.96) & 29.04 (+0.70) & 30.02 (+1.68) & 29.25 (+0.91) & 29.06 (+0.72) \\
LGV & 63.13 (+38.91) & \textbf{46.38 (+22.16)} & \textbf{54.98 (+30.76)} & \textbf{56.15 (+31.93)} & 30.28 (+1.94) & 30.53 (+2.19) & 30.88 (+2.54) & 32.47 (+4.13) \\
OPS & 67.96 (+43.74) & 28.28 (+4.06) & 31.46 (+7.24) & 31.11 (+6.89) & \textbf{32.01 (+3.67)} & 33.09 (+4.75) & 35.11 (+6.77) & 35.46 (+7.12) \\
PGN & 65.66 (+41.44) & 30.56 (+6.34) & 33.74 (+9.52) & 32.44 (+8.22) & 30.22 (+1.88) & \textbf{34.62 (+6.28)} & 36.00 (+7.66) & 35.33 (+6.99) \\
SSAH & 25.86 (+1.64) & 24.22 (+0.00) & 24.04 (-0.18) & 23.98 (-0.24) & 29.80 (+1.46) & 29.79 (+1.45) & 28.19 (-0.15) & 28.17 (-0.17) \\
VNI-FGSM & 58.16 (+33.94) & 30.41 (+6.19) & 34.84 (+10.62) & 34.00 (+9.78) & 30.01 (+1.67) & 34.35 (+6.01) & \textbf{36.41 (+8.07)} & 36.05 (+7.71) \\
\midrule
Mean & 57.43 (+33.21) & 29.58 (+5.36) & 32.38 (+8.16) & 31.97 (+7.75) & 30.04 (+1.70) & 31.75 (+3.41) & 32.56 (+4.22) & 32.72 (+4.38) \\
\bottomrule
\end{tabular}
\end{adjustbox}
\end{table*}

\subsubsection{Observations}
\noindent Our main observations are the following:
\begin{enumerate}
    \item Attacks using undefended surrogates perform better against undefended target models than attacks using defended surrogates on average. For CIFAR-10, undefended surrogates achieve a nominal \textbf{49.20\%} higher accuracy degradation than the highest-performing defended surrogate on average. Similarly, for ImageNet, the difference is \textbf{25.05\%} on average.
    \item Attacks using defended surrogates perform better against defended target models than attacks using undefended surrogates on average. For CIFAR-10, the lowest-performing defended surrogate exhibits a \textbf{4.61\%} nominal average accuracy degradation increase over the undefended surrogates. Likewise, for ImageNet, the average difference is \textbf{1.71\%}.
\end{enumerate}

These observations hold when using the mean of the degradations across all attacks, regardless of which of the three defended surrogate models are used. While they also hold for most individual attacks, they do not for some others. For instance, BIA and SSAH perform better using an undefended surrogate model than a defended surrogate model against a defended target model on ImageNet. Interestingly, these are the worst-performing attacks across our experiments, and we do not observe this behavior for the better-performing attacks on both datasets. 

We focus specifically on the best-performing attacks against defended target models, as an attacker might, namely Admix, VNI-FGSM, OPS, and PGN. In this case, we find that using a defended surrogate greatly improves the attack's transferability against defended target models. In the best case, for VNI-FGSM on CIFAR10, we note a roughly $\mathbf{4\times}$ increase in accuracy degradation, going from 3.40\% to 17.19\%. We also observe the same kind of improvement on ImageNet, with accuracy degradation increasing $\mathbf{3.86\times}$, going from 1.66\% to 8.07\%. Overall, our results indicate that when the attacker matches the defender's defense decision, their attack's success rate improves. This finding implies an incentive for the attacker to know and match the defender's decision. It follows that previous evaluations severely underestimated the potency of transferable attacks against defended models by only utilizing undefended surrogates, with the best degradation with a defended surrogate reaching 17.19\%, $\mathbf{3.73\times}$ more than the best degradation with an undefended surrogate at $4.61\%$.

\subsubsection{Other notable observations}
As previously mentioned, our main observations hold for most of our attacks except BIA and SSAH. Both attacks' poor performance is unlikely to be caused by an implementation error, as we reuse the code from each paper's original repository with minimal changes.

Another notable observation in our results is LGV's performance against undefended models. LGV is the best attack against undefended models in all cases but one (ImageNet, undefended surrogate vs. undefended target model), especially when using defended surrogates to attack undefended models, where it greatly outperforms other attacks. 
We believe this outperformance is due to LGV's unique attack process, where they continue to train the surrogate model with a high learning rate. This process overlooks any adversarial training or other robust training methods, which could potentially transform the defended surrogate model into a less robust one. This process would enable LGV to leverage its strong performance against undefended models, even when using a defended surrogate. 

Finally, we observe that the robustness of the surrogate model does not appear to impact attack performance against defended target models. While the worst-defended surrogate model underperforms in the aggregate, we find it to be the highest-performing surrogate for LGV, CDTP, BIA, and SSAH on CIFAR10. We also do not find a conclusive ranking between the median and best defended surrogates, with the median-defended surrogate outperforming on CIFAR10, but losing on ImageNet. We study this observation further with additional experimentation in Section \ref{subsec:surrogate_importance}.

\subsection{Importance of the robustness of the surrogate model} \label{subsec:surrogate_importance}
When conducting a transferable attack, the attacker must select a specific surrogate model. For defended surrogates, we present different options in Table \ref{tab:main-results-cifar10-w-base} and Table \ref{tab:main-results-imagenet-w-base}, namely the worst, median, and best defended surrogate models with respect to their performance against white-box AutoAttack. In this section, we investigate whether the robustness of the defended surrogate against white-box AutoAttack correlates with the transferability of the adversarial examples generated. 

\begin{figure}
    \centering
    \includegraphics[width=0.99\linewidth]{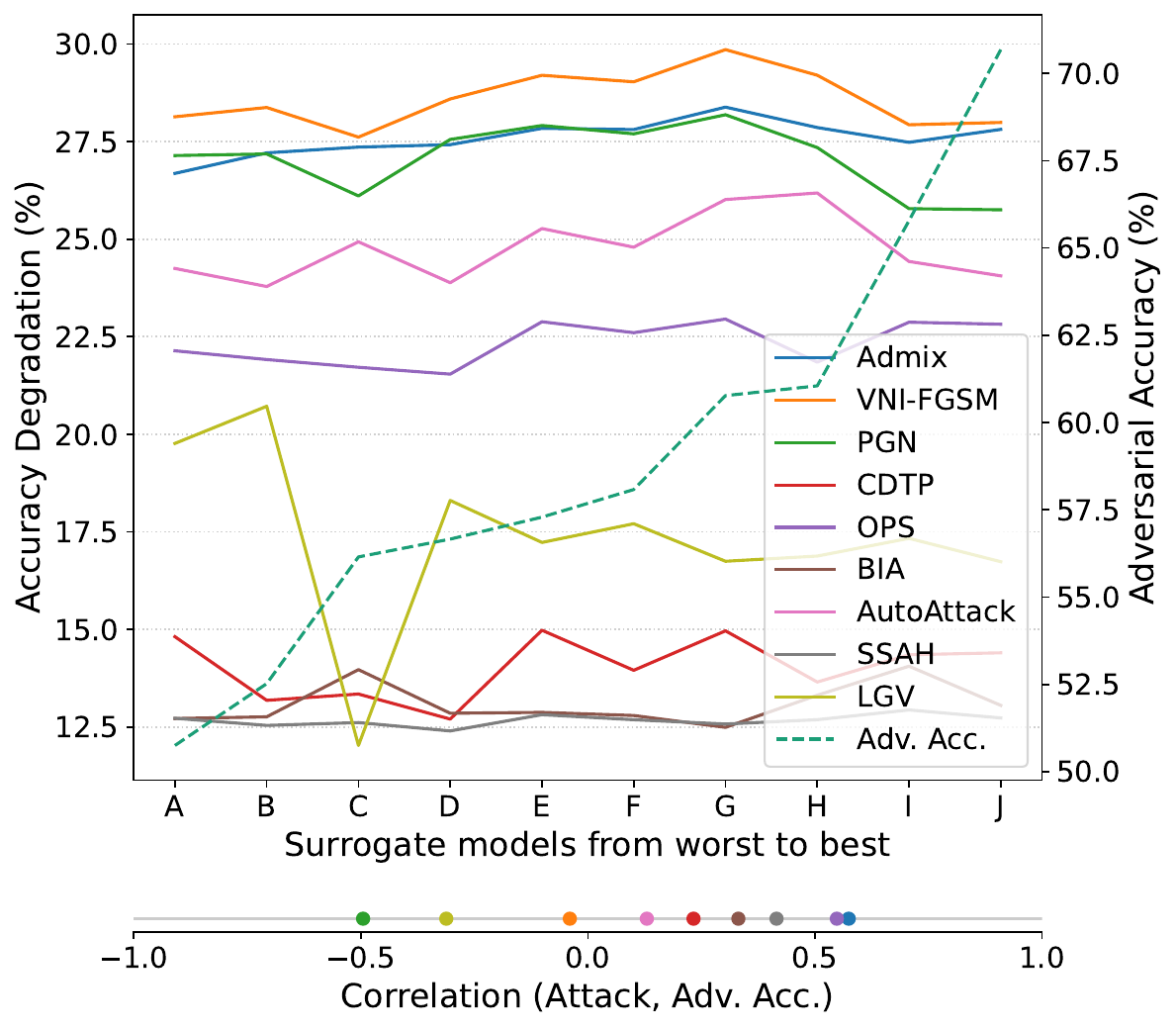}
    \caption{Transferability of adversarial examples from surrogate models of varying robustness. Models A–J are random representatives from successive deciles of defended CIFAR-10 models ranked by white-box AutoAttack adversarial accuracy (A = worst, J = best). Each surrogate is used to attack all other defended models with nine attacks. We also include the correlation between each attack and the adversarial accuracy below the main plot.}
    \label{fig:varying_surrogate}
\end{figure}

We rank the 68 defended models on CIFAR10 by adversarial accuracy against white-box AutoAttack, excluding the surrogate models used for Table \ref{tab:main-results-cifar10-w-base} and Table \ref{tab:main-results-imagenet-w-base}. Then, we partition the ranked list of models into deciles (0th-10th percentile, 10th-20th percentile, etc.). From each decile, we randomly select a model, yielding one representative per tranche. For simplicity, we label these models A through J, with A representing the 0-10th percentile tranche and J representing the 90-100th percentile tranche. For each representative, we attack all other CIFAR10 defended models with each of our nine attacks, and we plot the results in Figure \ref{fig:varying_surrogate}.
We also include each model's adversarial accuracy against white-box AutoAttack, as well as the correlation coefficient of each attack with this adversarial accuracy curve. Our results suggest that surrogate robustness is not correlated with better transferability, as indicated by the correlation coefficients ranging from -0.5 to 0.5. Thus, for the attacks we evaluate, the robustness of the surrogate model against white-box AutoAttack is not a significant factor in determining transferability. For that reason, we use the median-defended surrogate model as the default defended surrogate model for our analysis in Section \ref{sec:game-theory}. 

% {\color{red}[Should we add here the observation that the quality of the surrogate might not really matter?]}

\section{Game-theoretic analysis}\label{sec:game-theory}
While the general dilemma between transparency and security in AI systems is challenging to capture, we can examine an instance of this dilemma in transferable adversarial example attacks, using the conclusions of this study as an indicator for the general dilemma. In Section \ref{subsec:evaluation}, we present our main experimental results and find that matching the defender's decision can improve the attacker's success. This result suggests that transparency regarding the defender's decision to defend their model could conflict with the AI system's security against transferable adversarial example attacks. To model the trade-off between the transparency of the defender's decision and the model robustness,
 we need different games than the adversarial example game presented in Section \ref{subsec:adv_ex_attack_back}. The first of these new games, which we call the Surrogate game, is a static zero-sum two-player Nash game that we define as follows:\\
\noindent \textbf{Players:} The attacker is the row player and the defender is the column player. The attacker is associated with a fixed attack $A$. \\
\noindent \textbf{Actions:} The attacker chooses a distribution $\mathcal{S} \in \{\mathcal{U}, \mathcal{D}\}$ from the set containing the distribution of undefended models $\mathcal{U}$ and the distribution of defended models $\mathcal{D}$. The defender also chooses a distribution $\mathcal{T} \in \{\mathcal{U}, \mathcal{D}\}$.\\
\noindent \textbf{Utility functions:} The utility function of the defender is the expected adversarial accuracy of the target model's distribution:
\begin{equation}
     u_d(\mathcal{S}, \mathcal{T}) = \mathbb{E}_{(S, T, x,y) \sim \mathcal{S}\times \mathcal{T}\times \mathcal{Z}}[\mathbbm{1}(T(A(S, x, y)), y)]
\end{equation}

Since our game is zero-sum, the utility function of the attacker is $u_a(\mathcal{S}, \mathcal{T}) = 1 - u_d(\mathcal{S}, \mathcal{T})$ (accuracy degradation). 
In practice, we use the undefended and defended models we assemble, as described in Section \ref{subsec:setup}, to model $\mathcal{U}$ and $\mathcal{D}$. By default, when we refer to the expected payoff, we mean the expected payoff of the row player \cite{nisan2007vv}, which corresponds to the degradation in the target model's adversarial accuracy.

One aspect that the Surrogate game fails to account for is the defender's transparency concerning its decision. Since the Surrogate game is a Nash game, both players choose simultaneously. However, in the case where the defender is transparent about their decision, the attacker chooses after the defender, knowing the defender's choice. We model this second new game, where the defender is transparent, as a Stackelberg game with the same players, actions, and utility function as the Surrogate game, with the defender as the leader and the attacker as the follower. In this setting, the leader can only make pure commitments. By comparing the expected payoff of the Nash and Stackelberg games, we directly measure the security cost of transparency in terms of the degradation of adversarial accuracy.

\subsection{The Trade-off between Transparency and Security}\label{subsec:tradeoff}
\begin{table}[t]
\centering
\caption{Normal-form of the Surrogate game for VNI-FGSM on CIFAR10. We show the attacker's payoff (row player, in \%) since the game is zero-sum.}
\label{tab:vnifgsm-normal-form-cifar}
\begin{tabular}{lcc}
\toprule
 & \multicolumn{2}{c}{Defender's choice} \\
Attacker's choice & Undefended & Defended \\
\midrule
Undefended & 85.55 & 15.68 \\
Defended & 29.82 & 29.47 \\
\bottomrule
\end{tabular}
\end{table}
\begin{table}[t]
\centering
\caption{Normal-form of the Surrogate game for VNI-FGSM on ImageNet. We show the attacker's payoff (row player, in \%) since the game is zero-sum.}
\label{tab:vnifgsm-normal-form-imagenet}
\begin{tabular}{lcc}
\toprule
 & \multicolumn{2}{c}{Defender's choice} \\
Attacker's choice & Undefended & Defended \\
\midrule
Undefended & 58.16 & 30.01 \\
Defended & 34.84 & 36.41 \\
\bottomrule
\end{tabular}
\end{table}

Using our results from Table \ref{tab:main-results-cifar10-w-base} and Table \ref{tab:main-results-imagenet-w-base}, we create instances of the Surrogate Nash game for each attack and for each dataset. We then find the Nash equilibria for each instance of the game using the vertex enumeration method \cite{nisan2007vv} and compute the associated expected payoffs. We also repeat this process for the Surrogate Stackelberg game and compare the respective expected payoffs of the two games. If the two are equal, then transparency comes at no additional cost to the defender. However, when the expected payoff of the Stackelberg game is higher than that of the Nash game, it means that transparency lowers adversarial accuracy, as the attacker leverages the defender's transparency. Since the leader can only make pure commitments in our Stackelberg game, the expected payoff of the Stackelberg game will never be lower than that of the Nash game.

To illustrate, we select the VNI-FGSM attack and present the normal form of each Surrogate game for CIFAR10 and ImageNet in Table \ref{tab:vnifgsm-normal-form-cifar} and Table \ref{tab:vnifgsm-normal-form-imagenet}, respectively. We only show the attacker's payoff (row player) since the game is zero-sum and the attacker's payoff is $u_a = 1 - u_d$. 

On CIFAR10, we find a pure Nash equilibrium, where both the attacker and the defender choose to use a defended model. Under this Nash equilibrium, the attacker's payoff is 0.2947, meaning an accuracy degradation of 29.47\%, or a model adversarial accuracy of $100-29.47=70.53\%$. However, on ImageNet, we find a mixed Nash equilibrium, with an expected payoff for the attacker of 0.3607, corresponding to a 36.07\% accuracy degradation. Under this mixed Nash equilibrium, the probabilities of each action (Undefended, Defended) for the attacker are (0.0528, 0.9472), and the probabilities for the defender are (0.2153, 0.7847). This mixed equilibrium implies that, to maximize their overall adversarial accuracy, the defender should use an undefended model 21.53\% of the time. 

In the Surrogate Stackelberg game, we find that the expected payoff is the same as that of the Surrogate Nash game on CIFAR10. Yet, for ImageNet, we observe an increase in the attacker's payoff of 0.34\% compared to the Surrogate Nash game, since the defender can no longer use an undefended model 21.53\% of the time and must make a pure commitment. As a result, for VNI-FGSM on ImageNet, the defender's transparency about its defense decision harms security.

We repeat this comparative analysis for all of our attacks and report the resulting differences in expected attacker payoffs in Table \ref{tab:nash_stackelberg_diff_surrogate}. We find that for a total of 11 out of 18 attack-dataset pairs, transparency further reduces adversarial accuracy by 0.05\% to 1.04\% on CIFAR10 and between 0.02\% and 0.64\% on ImageNet. Considering the attacker only knows whether the model is defended, without any additional information about the target model or the defense itself, the observed reduction in adversarial accuracy supports the general notion that transparency can non-trivially damage model robustness.

\begin{table}[t!]
\centering
\caption{Difference (in \%) between the Nash game and the Stackelberg game expected attacker payoffs (Stackelberg - Nash). Bottom rows report the number of cases where payoffs differ and the mean change over nonzero entries.}
\label{tab:nash_stackelberg_diff_surrogate}
\begin{tabular}{lcc}
\toprule
Attack & CIFAR10 & ImageNet \\
\midrule
Admix & \textbf{0.28} & \textbf{0.32} \\
AutoAttack & \textbf{1.04} & \textbf{0.64} \\
BIA & 0 & 0 \\
CDTP & \textbf{0.05} & \textbf{0.02} \\
LGV & 0 & 0 \\
OPS & \textbf{0.49} & \textbf{0.29} \\
PGN & \textbf{0.42} & \textbf{0.35} \\
SSAH & 0 & 0 \\
VNI-FGSM & 0 & \textbf{0.34} \\
\midrule
\# Worse Off & 5 / 9 & 6 / 9 \\
Mean change ($\neq$ 0) & 0.46 & 0.33 \\
\bottomrule
\end{tabular}

\end{table}

In practice, an attacker would not be confined to using a singular attack method. Instead, they would select the attack with the best expected payoff among the available attacks. This selection changes the attacker's decision from choosing a surrogate distribution $\mathcal{S} \in \{\mathcal{U}, \mathcal{D}\}$ to choosing both an attack and a surrogate distribution $(A, \mathcal{S}) \in \mathcal{A} \times \{\mathcal{U}, \mathcal{D}\}$, where $\mathcal{A}$ is the set of all transferable adversarial attacks. We name this new game the Attack and Surrogate game (abbreviated as A\&S game). While the attacker's decision space changed to include the attack used, the defender's options remain the same: choosing between an undefended or defended target model. Likewise, the utility function for the defender remains the expected adversarial accuracy:
\begin{equation}
     u_d((A, \mathcal{S}), \mathcal{T}) = \mathbb{E}_{(A, S, T, x,y) \sim \mathcal{A} \times \mathcal{S}\times \mathcal{T}\times \mathcal{Z}}[\mathbbm{1}(T(A(S, x, y)), y)]
\end{equation}
and for the attacker, the expected accuracy degradation $u_a((A, \mathcal{S}), \mathcal{T}) = 1 - u_d((A, \mathcal{S}), \mathcal{T})$. The A\&S game is also a zero-sum game, like the Surrogate game.

\subsubsection{Attack and Surrogate Game}\label{subsec:attack_and_surrogate_game_analysis}
Similarly to the Surrogate game, we compare the A\&S Nash and Stackelberg games to measure the impact of transparency on adversarial accuracy. On CIFAR10, both games have the same optimal strategy for each player, where the attacker uses the VNI-FGSM attack with a defended surrogate, and the defender uses a defended target model. Hence, the expected payoffs of the Nash and Stackelberg equilibria are the same. However, on ImageNet, the expected payoffs differ by 0.18\%, meaning that, in the A\&S game, the defender's transparency about its defense decision can also harm security. When the defender is not transparent, following the Nash equilibrium strategy leads to the attacker using the OPS attack with an undefended surrogate with a 4.18\% probability and the VNI-FGSM attack with a defended surrogate with a 95.82\% probability. As for the defender, they should use an undefended target model 11.73\% of the time and a defended target model 88.27\% of the time.

\subsection{Implications of Optimal Play}
\begin{table}[t]
\centering
\caption{Probability (in \%) of using an undefended surrogate and target model for the attacker and defender, respectively, in the Surrogate game mixed Nash equilibria.}
\label{tab:prob_undefended}
\begin{adjustbox}{width=0.95\columnwidth}
\begin{tabular}{lcccc}
\toprule
 & \multicolumn{2}{c}{CIFAR10} & \multicolumn{2}{c}{ImageNet} \\
Attack & Surrogate & Target & Surrogate &  Target\\
\midrule
Admix & 2.45 & 13.84 & 5.87 & 12.66 \\
AutoAttack & 9.32 & 12.26 & 16.72 & 14.52 \\
CDTP & 8.49 & 0.83 & 9.60 & 0.56 \\
OPS & 8.86 & 7.10 & 9.22 & 7.83 \\
PGN & 3.41 & 14.94 & 5.99 & 15.33 \\
VNI-FGSM & 0.00 & 0.00 & 5.28 & 21.53 \\
\bottomrule
\end{tabular}

\end{adjustbox}
\end{table}

Our findings in Section \ref{subsec:tradeoff} show that transparency can be at odds with security in terms of model robustness against transferable adversarial example attacks. In the real world, defending or not defending are not the only options an agent has. While possible alternative actions could be more complex than simply using an undefended model, we find that even in this simplified scenario, obscurity improves defense performance. By being transparent about its decision and adhering to a pure strategy, the defender forfeits the option to choose alternative actions, which can compromise its robustness. Therefore, we study the instances of our games where the Nash equilibria are mixed, as they indicate scenarios where having more than one action available is beneficial to the defender.

In Table \ref{tab:prob_undefended}, we present for each dataset the probability with which an undefended model should be used by a player, with the Surrogate column representing how often an attacker should use an undefended surrogate and the Target column representing how often the defender should use an undefended target model. We only include the attacks with at least one mixed Nash equilibrium on either dataset. We find that, depending on the attack and the dataset, the defender should use an undefended target model with probabilities ranging from 0.83\% to 21.53\%. Likewise, since the best response for the attacker is to match the defender's decision, the attacker should use an undefended surrogate with probabilities ranging from 2.45\% to 16.72\%. When we consider the A\&S game rather than the Surrogate game, our analysis derives that the defender should use an undefended target model 11.73\% of the time. In contrast, the attacker should use an undefended surrogate with a 4.18\% probability. 

The existence of these mixed equilibria confirms that a pure defense strategy might not always be the best strategy for the defender. Given that this is the case for our scenario, which is a simplification of the real world, it suggests that this trade-off between defense transparency and security would also exist in more complex real-world scenarios.
We hypothesize that additional alternative actions for the defenders, such as other orthogonal defenses, when used as part of a mixed strategy, may offer much better improvements. However, this is outside the scope of this study. 

\subsection{The underestimation of the potency of transferable adversarial example attacks}
Existing benchmarks and evaluations of transferable adversarial example attacks have greatly underestimated their potency by only using undefended surrogates. To quantify the magnitude of this underestimation, we compare the expected payoffs of our A\&S game, where the attacker can choose both the attack it uses and whether its surrogate is defended or not, with a new game that we call the Attack game. The Attack game is the same as the A\&S game, except that the attacker must use an undefended surrogate model and can only choose which attack to use. By comparing the expected payoffs of the Attack and A\&S game, we empirically estimate the underestimation of the potency of transferable adversarial example attacks. We find a difference in expected accuracy degradation between the two games, amounting to \textbf{12.58\%} and \textbf{4.22\%} for CIFAR10 and ImageNet, respectively. These result in a $\mathbf{3.73\times}$ and a $\mathbf{2.15\times}$ increase in expected adversarial degradation when the attacker is not limited to using an undefended surrogate, meaning that transferable adversarial example attacks could be more successful against defended models than previously estimated.

\section{Conclusion}\label{sec:conclusion}
In this work, we investigate the trade-off between transparency and security in adversarial machine learning through the lens of transferable adversarial example attacks. Our large-scale study on CIFAR10 and ImageNet showed that the attacker can benefit from matching the defender's defense decision. Thus, revealing whether the defender defended their model can worsen their security. Our study also highlights an underestimation of the potency of transferable attacks against defended models by previous work.

We formalize these interactions as Nash and Stackelberg games, depending on whether the defender is transparent or not. Then, we demonstrate that transparency, as prescribed by Responsible AI, can incur a security cost. This cost occurs even for something as minor as revealing that the target model is defended, without providing any additional information about the target model or the defense itself. While security through obscurity is often considered insufficient on its own, our results show that, in the case of transferable adversarial example attacks, obscuring the defense status is a beneficial strategy for the defender.

Finally, our results suggest that pure and overt strategies are unlikely to lead to optimal protection. Instead, they indicate that diversifying defenses as part of an overall defense strategy could enhance the chances of success for the defender, especially when these defenses do not transfer well between one another. Further work into combining orthogonal defenses could improve our understanding of transferable adversarial example attacks and defenses.
Additionally, future work could extend our analysis to other modalities beyond image classification and other types of attacks. In a broader context, our results and analysis suggest that transparency practices in Responsible AI can potentially introduce unintended vulnerabilities. Understanding where transparency strengthens trust and where it undermines security is an important direction for Responsible AI research.

% \begin{acks}
% If you wish to include any acknowledgments in your paper (e.g., to 
% people or funding agencies), please do so using the `\texttt{acks}' 
% environment. Note that the text of your acknowledgments will be omitted
% if you compile your document with the `\texttt{anonymous}' option.
% \end{acks}

%%%%%%%%%%%%%%%%%%%%%%%%%%%%%%%%%%%%%%%%%%%%%%%%%%%%%%%%%%%%%%%%%%%%%%%%

%%% The next two lines define, first, the bibliography style to be 
%%% applied, and, second, the bibliography file to be used.

\bibliographystyle{ACM-Reference-Format} 
\bibliography{main}

@misc{cifar10,
	author= "Alex Krizhevsky",
	title= "Learning multiple layers of features from tiny images",
	note= "In \textit{Technical report}",
	year= "2009",
}

@inproceedings{Imagenet,
  title={Imagenet: A large-scale hierarchical image database},
  author={Deng, Jia and Dong, Wei and Socher, Richard and Li, Li-Jia and Li, Kai and Fei-Fei, Li},
  booktitle={2009 IEEE conference on computer vision and pattern recognition},
  pages={248--255},
  year={2009},
  organization={Ieee}
}

@article{robustbench,
  title={Robustbench: a standardized adversarial robustness benchmark},
  author={Croce, Francesco and Andriushchenko, Maksym and Sehwag, Vikash and Debenedetti, Edoardo and Flammarion, Nicolas and Chiang, Mung and Mittal, Prateek and Hein, Matthias},
  journal={arXiv preprint arXiv:2010.09670},
  year={2020}
}

@misc{pytorch-cifar,
    key = {Pytorch-cifar},
    title = {Train CIFAR10 with PyTorch},
    url = {https://github.com/kuangliu/pytorch-cifar},
}

@article{Pytorch,
  title={Automatic differentiation in PyTorch},
  author={Paszke, Adam and Gross, Sam and Chintala, Soumith and Chanan, Gregory and Yang, Edward and DeVito, Zachary and Lin, Zeming and Desmaison, Alban and Antiga, Luca and Lerer, Adam},
  year={2017}
}

@InProceedings{resnet,
author = {He, Kaiming and Zhang, Xiangyu and Ren, Shaoqing and Sun, Jian},
title = {Deep Residual Learning for Image Recognition},
booktitle = {Proceedings of the IEEE Conference on Computer Vision and Pattern Recognition (CVPR)},
month = {June},
year = {2016}
}

@inproceedings{autopgd,
  title={Reliable evaluation of adversarial robustness with an ensemble of diverse parameter-free attacks},
  author={Croce, Francesco and Hein, Matthias},
  booktitle={International conference on machine learning},
  pages={2206--2216},
  year={2020},
  organization={PMLR}
}

@article{vassilev2024adversarial,
  title={Adversarial machine learning: A taxonomy and terminology of attacks and mitigations},
  author={Vassilev, Apostol and Oprea, Alina and Fordyce, Alie and Andersen, Hyrum},
  year={2024},
}

@inproceedings{lecuyer2019certified,
  title={Certified robustness to adversarial examples with differential privacy},
  author={Lecuyer, Mathias and Atlidakis, Vaggelis and Geambasu, Roxana and Hsu, Daniel and Jana, Suman},
  booktitle={2019 IEEE symposium on security and privacy (SP)},
  pages={656--672},
  year={2019},
  organization={IEEE}
}

@inproceedings{katz2017reluplex,
  title={Reluplex: An efficient SMT solver for verifying deep neural networks},
  author={Katz, Guy and Barrett, Clark and Dill, David L and Julian, Kyle and Kochenderfer, Mykel J},
  booktitle={International conference on computer aided verification},
  pages={97--117},
  year={2017},
  organization={Springer}
}

@article{madry2017towards,
  title={Towards deep learning models resistant to adversarial attacks},
  author={Madry, Aleksander and Makelov, Aleksandar and Schmidt, Ludwig and Tsipras, Dimitris and Vladu, Adrian},
  journal={arXiv preprint arXiv:1706.06083},
  year={2017}
}

@article{wu2023defenses,
  title={Defenses in adversarial machine learning: A survey},
  author={Wu, Baoyuan and Wei, Shaokui and Zhu, Mingli and Zheng, Meixi and Zhu, Zihao and Zhang, Mingda and Chen, Hongrui and Yuan, Danni and Liu, Li and Liu, Qingshan},
  journal={arXiv preprint arXiv:2312.08890},
  year={2023}
}

@article{dasgupta2019survey,
  title={A survey of game theoretic approaches for adversarial machine learning in cybersecurity tasks},
  author={Dasgupta, Prithviraj and Collins, Joseph},
  journal={AI Magazine},
  volume={40},
  number={2},
  pages={31--43},
  year={2019}
}

@article{li2024survey,
  title={A survey of decision making in adversarial games},
  author={Li, Xiuxian and Meng, Min and Hong, Yiguang and Chen, Jie},
  journal={Science China Information Sciences},
  volume={67},
  number={141201},
  year={2024},
  publisher={Springer}
}

@article{kurakin2016adversarial,
  title={Adversarial machine learning at scale},
  author={Kurakin, Alexey and Goodfellow, Ian and Bengio, Samy},
  journal={arXiv preprint arXiv:1611.01236},
  year={2016}
}

@article{zhao2023revisiting,
  title={Revisiting transferable adversarial image examples: Attack categorization, evaluation guidelines, and new insights},
  author={Zhao, Zhengyu and Zhang, Hanwei and Li, Renjue and Sicre, Ronan and Amsaleg, Laurent and Backes, Michael and Li, Qi and Shen, Chao},
  journal={arXiv preprint arXiv:2310.11850},
  year={2023}
}

@article{rathbun2022game,
  title={Game theoretic mixed experts for combinational adversarial machine learning},
  author={Rathbun, Ethan and Mahmood, Kaleel and Ahmad, Sohaib and Ding, Caiwen and Van Dijk, Marten},
  journal={arXiv preprint arXiv:2211.14669},
  year={2022}
}

@article{do2017game,
  title={Game theory for cyber security and privacy},
  author={Do, Cuong T and Tran, Nguyen H and Hong, Choongseon and Kamhoua, Charles A and Kwiat, Kevin A and Blasch, Erik and Ren, Shaolei and Pissinou, Niki and Iyengar, Sundaraja Sitharama},
  journal={ACM Computing Surveys (CSUR)},
  volume={50},
  number={2},
  pages={1--37},
  year={2017},
  publisher={ACM New York, NY, USA}
}

@article{manshaei2013game,
  title={Game theory meets network security and privacy},
  author={Manshaei, Mohammad Hossein and Zhu, Quanyan and Alpcan, Tansu and Bac{\c{s}}ar, Tamer and Hubaux, Jean-Pierre},
  journal={Acm Computing Surveys (Csur)},
  volume={45},
  number={3},
  pages={1--39},
  year={2013},
  publisher={ACM New York, NY, USA}
}

@article{pawlick2019game,
  title={A game-theoretic taxonomy and survey of defensive deception for cybersecurity and privacy},
  author={Pawlick, Jeffrey and Colbert, Edward and Zhu, Quanyan},
  journal={ACM Computing Surveys (CSUR)},
  volume={52},
  number={4},
  pages={1--28},
  year={2019},
  publisher={ACM New York, NY, USA}
}

@article{ml_healthcare,
    author = {Hafsa Habehhl and Suril Gohel},
    title = {Machine Learning in Healthcare},
    journal = {national library of medicine},
    year = {2021},
}

@article{GUPTA2021100057,
title = {Deep learning for object detection and scene perception in self-driving cars: Survey, challenges, and open issues},
journal = {Array},
volume = {10},
pages = {100057},
year = {2021},
author = {Abhishek Gupta and Alagan Anpalagan and Ling Guan and Ahmed Shaharyar Khwaja}
}

@article{Ni_2022,
year = {2022},
month = {dec},
publisher = {IOP Publishing},
volume = {2386},
number = {1},
pages = {012031},
author = {Zhehan Ni},
title = {Reframe the Field of Aerospace Engineering Via Machine Learning: Application and Comparison},
journal = {Journal of Physics: Conference Series}
}

@article{stackelberg1934marktform,
  title={Marktform und gleichgewicht},
  author={Stackelberg, Heinrich von},
  journal={(No Title)},
  year={1934}
}

@inproceedings{araujo2020advocating,
  title={Advocating for multiple defense strategies against adversarial examples},
  author={Araujo, Alexandre and Meunier, Laurent and Pinot, Rafael and Negrevergne, Benjamin},
  booktitle={Joint European Conference on Machine Learning and Knowledge Discovery in Databases},
  pages={165--177},
  year={2020},
  organization={Springer}
}

@article{pal2020game,
  title={A game theoretic analysis of additive adversarial attacks and defenses},
  author={Pal, Ambar and Vidal, Ren{\'e}},
  journal={Advances in Neural Information Processing Systems},
  volume={33},
  pages={1345--1355},
  year={2020}
}

@inproceedings{sengupta2019mtdeep,
  title={Mtdeep: boosting the security of deep neural nets against adversarial attacks with moving target defense},
  author={Sengupta, Sailik and Chakraborti, Tathagata and Kambhampati, Subbarao},
  booktitle={International Conference on Decision and Game Theory for Security},
  pages={479--491},
  year={2019},
  organization={Springer}
}

@article{bose2020adversarial,
  title={Adversarial example games},
  author={Bose, Joey and Gidel, Gauthier and Berard, Hugo and Cianflone, Andre and Vincent, Pascal and Lacoste-Julien, Simon and Hamilton, Will},
  journal={Advances in neural information processing systems},
  volume={33},
  pages={8921--8934},
  year={2020}
}

@article{xu2017feature,
  title={Feature squeezing: Detecting adversarial examples in deep neural networks},
  author={Xu, Weilin and Evans, David and Qi, Yanjun},
  journal={arXiv preprint arXiv:1704.01155},
  year={2017}
}

@article{nie2022diffusion,
  title={Diffusion models for adversarial purification},
  author={Nie, Weili and Guo, Brandon and Huang, Yujia and Xiao, Chaowei and Vahdat, Arash and Anandkumar, Anima},
  journal={arXiv preprint arXiv:2205.07460},
  year={2022}
}

@article{shi2021online,
  title={Online adversarial purification based on self-supervision},
  author={Shi, Changhao and Holtz, Chester and Mishne, Gal},
  journal={arXiv preprint arXiv:2101.09387},
  year={2021}
}

@article{na2017cascade,
  title={Cascade adversarial machine learning regularized with a unified embedding},
  author={Na, Taesik and Ko, Jong Hwan and Mukhopadhyay, Saibal},
  journal={arXiv preprint arXiv:1708.02582},
  year={2017}
}

@misc{mnist,
	author= "Yann LeCun",
	title= "The mnist database of handwritten digits",
	note= "In \textit{Technical report}",
	year= "1998",
}

@inproceedings{wang2021admix,
  title={Admix: Enhancing the transferability of adversarial attacks},
  author={Wang, Xiaosen and He, Xuanran and Wang, Jingdong and He, Kun},
  booktitle={Proceedings of the IEEE/CVF international conference on computer vision},
  pages={16158--16167},
  year={2021}
}

@inproceedings{wang2021enhancing,
  title={Enhancing the transferability of adversarial attacks through variance tuning},
  author={Wang, Xiaosen and He, Kun},
  booktitle={Proceedings of the IEEE/CVF conference on computer vision and pattern recognition},
  pages={1924--1933},
  year={2021}
}

@inproceedings{gubri2022lgv,
  title={Lgv: Boosting adversarial example transferability from large geometric vicinity},
  author={Gubri, Martin and Cordy, Maxime and Papadakis, Mike and Traon, Yves Le and Sen, Koushik},
  booktitle={European Conference on Computer Vision},
  pages={603--618},
  year={2022},
  organization={Springer}
}

@inproceedings{luo2022frequency,
  title={Frequency-driven imperceptible adversarial attack on semantic similarity},
  author={Luo, Cheng and Lin, Qinliang and Xie, Weicheng and Wu, Bizhu and Xie, Jinheng and Shen, Linlin},
  booktitle={Proceedings of the IEEE/CVF conference on computer vision and pattern recognition},
  pages={15315--15324},
  year={2022}
}

@article{zhang2022beyond,
  title={Beyond imagenet attack: Towards crafting adversarial examples for black-box domains},
  author={Zhang, Qilong and Li, Xiaodan and Chen, Yuefeng and Song, Jingkuan and Gao, Lianli and He, Yuan and Xue, Hui},
  journal={arXiv preprint arXiv:2201.11528},
  year={2022}
}

@inproceedings{guo2025boosting,
  title={Boosting Adversarial Transferability through Augmentation in Hypothesis Space},
  author={Guo, Yu and Liu, Weiquan and Xu, Qingshan and Zheng, Shijun and Huang, Shujun and Zang, Yu and Shen, Siqi and Wen, Chenglu and Wang, Cheng},
  booktitle={Proceedings of the Computer Vision and Pattern Recognition Conference},
  pages={19175--19185},
  year={2025}
}

@article{ge2023boosting,
  title={Boosting adversarial transferability by achieving flat local maxima},
  author={Ge, Zhijin and Liu, Hongying and Xiaosen, Wang and Shang, Fanhua and Liu, Yuanyuan},
  journal={Advances in Neural Information Processing Systems},
  volume={36},
  pages={70141--70161},
  year={2023}
}

@article{naseer2019cross,
  title={Cross-domain transferability of adversarial perturbations},
  author={Naseer, Muhammad Muzammal and Khan, Salman H and Khan, Muhammad Haris and Shahbaz Khan, Fahad and Porikli, Fatih},
  journal={Advances in Neural Information Processing Systems},
  volume={32},
  year={2019}
}

@inproceedings{papernot2018sok,
  title={Sok: Security and privacy in machine learning},
  author={Papernot, Nicolas and McDaniel, Patrick and Sinha, Arunesh and Wellman, Michael P},
  booktitle={2018 IEEE European symposium on security and privacy (EuroS\&P)},
  pages={399--414},
  year={2018},
  organization={IEEE}
}

@article{batool2023responsible,
  title={Responsible AI governance: a systematic literature review},
  author={Batool, Amna and Zowghi, Didar and Bano, Muneera},
  journal={arXiv preprint arXiv:2401.10896},
  year={2023}
}

@misc{transferattack,
    key = {TransferAttack},
    title = {Devling into Adversarial Transferability on Image Classification: A Review, Benchmark and Evaluation},
    url = {https://github.com/Trustworthy-AI-Group/TransferAttack},
}

@misc{nisan2007vv,
  title={VV (Eds.), Algorithmic Game Theory},
  author={Nisan, Noam and Roughgarden, Tim and Tardos, Eva},
  year={2007},
  publisher={Cambridge University Press}
}

@inproceedings{wang2023towards,
  title={Towards transferable targeted adversarial examples},
  author={Wang, Zhibo and Yang, Hongshan and Feng, Yunhe and Sun, Peng and Guo, Hengchang and Zhang, Zhifei and Ren, Kui},
  booktitle={Proceedings of the IEEE/CVF conference on computer vision and pattern recognition},
  pages={20534--20543},
  year={2023}
}

@inproceedings{andriushchenko2020square,
  title={Square attack: a query-efficient black-box adversarial attack via random search},
  author={Andriushchenko, Maksym and Croce, Francesco and Flammarion, Nicolas and Hein, Matthias},
  booktitle={European conference on computer vision},
  pages={484--501},
  year={2020},
  organization={Springer}
}

@article{li2023towards,
  title={Towards evaluating transfer-based attacks systematically, practically, and fairly},
  author={Li, Qizhang and Guo, Yiwen and Zuo, Wangmeng and Chen, Hao},
  journal={Advances in Neural Information Processing Systems},
  volume={36},
  pages={41707--41726},
  year={2023}
}

@misc{anthropic2025constitutional,
  title        = {Constitutional Classifiers: Defending against universal jailbreaks},
  author       = {{Anthropic}},
  howpublished = {\url{https://www.anthropic.com/news/constitutional-classifiers}},
  year         = {2025},
}

@misc{anthropic2025claudechrome,
  title        = {Piloting Claude for Chrome},
  author       = {{Anthropic}},
  howpublished = {\url{https://www.anthropic.com/news/claude-for-chrome}},
  year         = {2025},
}

@misc{openai2025traderobustness,
  title        = {Trading Inference-Time Compute for Adversarial Robustness},
  author       = {{OpenAI}},
  howpublished = {\url{https://openai.com/index/trading-inference-time-compute-for-adversarial-robustness/}},
  year         = {2025},
}

@inproceedings{meunier2021mixed,
  title={Mixed nash equilibria in the adversarial examples game},
  author={Meunier, Laurent and Scetbon, Meyer and Pinot, Rafael B and Atif, Jamal and Chevaleyre, Yann},
  booktitle={International Conference on Machine Learning},
  pages={7677--7687},
  year={2021},
  organization={PMLR}
}

@incollection{kurakin2018adversarial,
  title={Adversarial examples in the physical world},
  author={Kurakin, Alexey and Goodfellow, Ian J and Bengio, Samy},
  booktitle={Artificial intelligence safety and security},
  pages={99--112},
  year={2018},
  publisher={Chapman and Hall/CRC}
}

%%%%%%%%%%%%%%%%%%%%%%%%%%%%%%%%%%%%%%%%%%%%%%%%%%%%%%%%%%%%%%%%%%%%%%%%

\end{document}